\documentclass[12pt]{article}

\usepackage[margin=1in]{geometry}

\usepackage{setspace}

\usepackage{graphicx}

\usepackage{enumitem}

\usepackage{microtype}
\usepackage{subfigure}
\usepackage{booktabs}
\usepackage{comment}
\usepackage{wrapfig}
\usepackage{arydshln}
\usepackage{enumitem}
\usepackage{multirow}
\usepackage{url}
\usepackage{natbib}
\usepackage{authblk}

\RequirePackage[colorlinks,citecolor=blue,linkcolor=blue,urlcolor=blue,pagebackref]{hyperref}

\usepackage{amsmath}
\usepackage{amssymb}
\usepackage{mathtools}
\usepackage{amsfonts}
\usepackage{bm}
\usepackage{bbm}

\usepackage{algorithm}
\usepackage{algorithmic}

\def\eqref#1{equation~\ref{#1}}

\def\1{\bm{1}}

\def\rmd{{\mathrm{d}}}

\DeclareMathAlphabet{\mathsfit}{\encodingdefault}{\sfdefault}{m}{sl}
\SetMathAlphabet{\mathsfit}{bold}{\encodingdefault}{\sfdefault}{bx}{n}

\def\calA{{\mathcal{A}}}

\def\calR{{\mathcal{R}}}

\def\calW{{\mathcal{W}}}
\def\calX{{\mathcal{X}}}
\def\calY{{\mathcal{Y}}}
\def\calZ{{\mathcal{Z}}}

\def\bbE{{\mathbb{E}}}

\def\bbN{{\mathbb{N}}}

\def\bbR{{\mathbb{R}}}

\DeclareMathOperator*{\argmin}{arg\,min}

\newcommand{\p}[1]{\left(#1\right)}
\newcommand{\sqb}[1]{\left[#1\right]}
\newcommand{\cb}[1]{\left\{#1\right\}}

\newcommand{\bigp}[1]{\big(#1\big)}

\newcommand{\Bigp}[1]{\Big(#1\Big)}

\usepackage{amsthm}

\theoremstyle{plain}

\usepackage[textsize=tiny]{todonotes}
\usepackage{multirow}
\usepackage{wrapfig}
\usepackage{subfigure}

\renewcommand{\eqref}[1]{(\ref{#1})}

\usepackage{xcolor}
\newcount\Comments  
\newcommand{\kibitz}[2]{\ifnum\Comments=1\textcolor{#1}{#2}\fi}

\usepackage[capitalize,noabbrev]{cleveref}

\allowdisplaybreaks

\title{Riesz Regression As Direct Density Ratio Estimation}

\author{Masahiro Kato\thanks{Email: \texttt{mkato-csecon@g.ecc.u-tokyo.ac.jp}}$\,$}

\affil{The University of Tokyo}

\date{\today}

\begin{document}

\maketitle

\begin{abstract}
This study clarifies the relationship between Riesz regression \citep{Chernozhukov2021automaticdebiased} and density ratio estimation (DRE) in causal inference problems, such as average treatment effect estimation. We first show that the Riesz representer can be written as a signed density ratio and then demonstrate that the Riesz regression objective coincides with the least-squares importance fitting criterion \citep{Kanamori2009aleastsquares}. Although Riesz regression applies to a broad class of representer estimation problems, this equivalence with DRE allows us to transfer existing DRE results, including convergence rate analyses, generalizations based on Bregman divergence minimization, and regularization techniques for flexible models such as neural networks.\end{abstract}

\section{Introduction}
We describe the connection between Riesz representer estimation in causal inference and density ratio estimation (DRE) \citep{Chernozhukov2022automaticdebiased,Sugiyama2012densityratio}. With the advance of various machine learning methods, debiased machine learning (DML) approaches garnered attention in economics and finance \citep{Movaghari2025coporatecash,Kumar2024unveilingthe,Shi2019adaptingneural}. The automatic DML (ADML) framework formulates causal and structural parameter estimation problems using the Riesz representer and provides Riesz regression, a general tool for Riesz representer estimation \citep{Chen2015sievesemiparametric,Chernozhukov2021automaticdebiased}. This paper shows that (i) the Riesz representer can be written as a signed density ratio (Section~\ref{sec:main1}) and (ii) the Riesz regression criterion in \citet{Chernozhukov2021automaticdebiased} is identical to the least-squares importance fitting (LSIF) criterion in \citet{Kanamori2009aleastsquares} for DRE (Section~\ref{sec:main2}). In discrete treatment problems, the first result yields closed-form Riesz representers for average treatment effect (ATE), average treatment effect on the treated (ATT), weighted ATE, policy value, and other linear contrasts. For simplicity, we specialize the discussion to the ATE, a canonical target in causal inference.

This observation is useful for applied work because the representer is the weighting component in Neyman orthogonal scores. Writing it as a density ratio turns direct representer fitting into direct reweighting. The same step links orthogonal score estimation to methods already familiar in causal inference, including tailored loss, entropy balancing, stable balancing weights, nearest neighbor matching, and augmented balancing weights \citep{Zhao2019covariatebalancing,Hainmueller2012entropybalancing,Zubizarreta2015stableweights,Lin2023estimationbased,BrunsSmith2025augmentedbalancing}. It also connects Riesz estimation to model classes and regularization tools developed for DRE, such as RKHS estimators, neural networks, nonnegative Bregman corrections, and telescoping DRE \citep{Kanamori2012statisticalanalysis,Kato2021nonnegativebregman,Zheng2022anerror,Rhodes2020telescopingdensityratio}. Related work develops this relationship for direct bias correction, Bregman-Riesz regression, nearest neighbor matching, and unified treatments of weighting and matching \citep{Kato2025directbias,Kato2025nearestneighbor,Kato2026unifiedframework}.

This focus on the Riesz representer also helps with interpretation. In empirical work, the outcome regression usually receives the most attention, while the representer is treated as a technical ingredient needed for debiasing. For treatment effect problems, however, the representer is the estimand-specific weighting rule. Once it is written as a density ratio, questions about balancing, overlap, and regularization become questions about how to estimate that weighting rule directly. This shift in viewpoint is the main reason the link to DRE is useful beyond a purely formal equivalence.

\section{Setup}
Let $X\in \calX$ be a regressor and $Y\in\calY \subseteq \bbR$ be a scalar outcome, where $\calX$ is the regressor space and $\calY$ is the outcome space. Let us assume that $W = (X, Y) \in \calW \coloneqq \calX \times \calY$ jointly follows a distribution $P$. Let $\{(X_i, Y_i)\}_{i=1}^n$ be observations, where the sample size is $n\in\bbN$ and each $(X_i, Y_i)$ is an i.i.d. copy from $P$. 

\paragraph{Parameter of interest.}
We denote the parameter of interest by $\theta_0$. Let $\gamma_0 \coloneq \bbE\sqb{Y\mid X}$ be the regression function, and let $\Gamma$ be the set of measurable functions $\gamma \colon \calX\to \bbR$ such that $\bbE\sqb{\gamma(X)^2} < \infty$. We assume that the parameter of interest $\theta_0$ can be written as
\[\theta_0 \coloneqq \bbE\sqb{m(W, \gamma_0)},\]
where $m\colon \calW \times \Gamma \to \bbR$ is such that $m(w,\cdot)$ is linear for each $w\in\calW$.

\paragraph{Goal.}
Our goal is to estimate $\theta_0$ at the $\sqrt{n}$ rate and with asymptotic efficiency, that is, asymptotic normality with asymptotic variance matching the efficiency bound \citep{Vaart1998asymptoticstatistics}.

\section{Riesz Representers and Density Ratios}
\label{sec:main1}
In efficient estimation of $\theta_0$, Neyman orthogonality plays an important role, and it can be expressed using the Riesz representer. In this section, we recap the Neyman orthogonal score and the Riesz representer and show that the Riesz representer can be written as a signed density ratio.

\subsection{Neyman Orthogonal Scores and Riesz Representer}
Neyman orthogonal scores matter because asymptotically linear estimators based on them are asymptotically efficient, and orthogonality reduces plug-in bias when nuisance functions are replaced by estimators.

When the parameter of interest $\theta_0$ is linear in the regression function, let
\[
L(\gamma) \coloneqq \bbE\sqb{m(W,\gamma)}.
\]
The Riesz representer $\alpha_0$ is the function satisfying
\[
L(\gamma)=\bbE\sqb{\alpha_0(X)\gamma(X)}
\]
for all $\gamma\in\Gamma$. The Neyman orthogonal score is then given by \citep{Newey1994theasymptotic,Chernozhukov2021automaticdebiased}
\[
\psi(W; \gamma_0, \alpha_0, \theta_0) \coloneqq \alpha_0(X)\bigp{Y-\gamma_0(X)} + m(W,\gamma_0) - \theta_0.
\]

In practice, both $\gamma_0$ and $\alpha_0$ are unknown and must be estimated. The role of $\alpha_0$ is easy to overlook because it is introduced through semiparametric geometry, but in treatment effect problems it is the object that determines how residuals are reweighted inside the orthogonal score. This paper exploits that interpretation.

\subsection{Signed Density Ratio}
Let $P_X$ be the law of the regressor $X$ under $P$. If a linear functional can be written as
\[
L(\gamma)=\int \gamma(x)d\nu(x),
\]
for a finite signed measure $\nu$ such that $\nu \ll P_X$ and $d\nu/dP_X \in L_2(P_X)$, then
\[
\alpha_0(x)=\frac{\rmd \nu}{\rmd P_X}(x)
\]
is the Riesz representer, because $L(\gamma)=\bbE\sqb{\alpha_0(X)\gamma(X)}$. Hence, the representer is a signed density ratio whenever the target is a signed change of measure.

This statement makes the Riesz representer concrete. It is not an abstract Hilbert space object but the derivative that tilts the law of the regression argument toward the target functional. The sign and magnitude of $\alpha_0$ describe how observations are reweighted to recover the parameter of interest. In treatment effect problems, overlap is precisely the condition that keeps this ratio stable.

A practically important class is discrete treatment with linear mean functionals. Let $D\in\cb{0,1,\dots,K}$, let $e_{0,d}(z)\coloneqq \Pr(D=d\mid Z=z)$, and consider
\[
\theta_0 \coloneqq \bbE\sqb{\sum^K_{d=0} h_d(Z)\gamma_0(d,Z)},
\]
where $\cb{h_d}_{d=0}^K$ are known weighting functions. Then the representer is
\[
\alpha_0(D,Z)\coloneqq \sum^K_{d=0} h_d(Z)\frac{\mathbbm{1}\sqb{D=d}}{e_{0,d}(Z)},
\]
since
\[
\bbE\sqb{\alpha_0(D,Z)\gamma(D,Z)}=\bbE\sqb{\sum^K_{d=0} h_d(Z)\gamma(d,Z)}.
\]
This class includes several targets used in practice. ATE corresponds to $K=1$ with $h_1(Z)=1$ and $h_0(Z)=-1$. Weighted ATE uses $h_1(Z)=w(Z)$ and $h_0(Z)=-w(Z)$. Policy value uses $h_d(Z)=\mathbbm{1}\sqb{\pi(Z)=d}$. For more details on DRE for ATE estimation, see Section~\ref{sec:equivrieszdr}.

ATT also belongs to the same class because
\[
\theta_0=\bbE\sqb{\frac{e_0(Z)}{p_1}\gamma_0(1,Z)-\frac{e_0(Z)}{p_1}\gamma_0(0,Z)},
\]
where $p_1\coloneqq \Pr(D=1)$, so that
\[
\alpha_0(D,Z)=\frac{D}{p_1}-\frac{e_0(Z)}{p_1\p{1-e_0(Z)}}(1-D).
\]

The same logic covers other targets that average $\gamma_0(a,Z)$ with respect to a target covariate distribution rather than the observed one. In such cases, the representer is again the appropriate change-of-measure ratio multiplied by the treatment indicator. This is exactly the type of problem for which direct DRE has been designed.

\section{ATE Estimation}
The remainder of the paper explains the equivalence between Riesz regression and LSIF, focusing on the ATE case, where the equivalence is clearest.

\subsection{Setup}
\paragraph{Potential outcomes and observations}
We consider a binary treatment, where $1$ denotes treatment and $0$ denotes control. Following the Neyman-Rubin causal framework \citep{Neyman1923surapplications,Rubin1974estimatingcausal}, we define potential outcomes $Y(1), Y(0) \in \calY$, where $\calY \subseteq \bbR$ denotes the outcome space. Let $Z \in \calZ$ be unit-level covariates, where $\calZ$ denotes the covariate space. Let $D\in\cb{0,1}$ be a treatment indicator, and let $Y$ be the observed outcome defined as
\[
Y = DY(1) + (1-D)Y(0).
\]
For the regressor $X = (D, Z)$, define
\[
\gamma_0(X) = \gamma_0(D, Z) \coloneqq \bbE\sqb{Y\mid D, Z}.
\]
We observe an i.i.d. sample
\[\cb{(D_i, Z_i, Y_i)}_{i=1}^n.\]

\paragraph{ATE}
Using the observations, we aim to estimate the ATE. The ATE is defined as
\[
\theta^{\text{ATE}}_0 \coloneqq \bbE\sqb{Y(1)-Y(0)}.
\]
Under unconfoundedness, it can also be written as $\theta^{\text{ATE}}_0 = \bbE\sqb{\gamma_0(1, Z) - \gamma_0(0, Z)}$.

\paragraph{Notation and assumptions}
Let $e_0(z)\coloneqq \Pr(D=1\mid Z=z)$ be the propensity score. We impose unconfoundedness and overlap, that is, $Y(1),Y(0)\perp D\mid Z$ and there exists $\epsilon\in(0,1/2)$ such that $\epsilon<e_0(Z)<1-\epsilon$ almost surely. We also write $p_{D,Z}(d,z)$ for the joint density of $(D,Z)$ and $p_Z(z)$ for the marginal density of $Z$ when they exist.

\subsection{Neyman Orthogonal Score and Riesz Representer in ATE Estimation}
In ATE estimation, the moment function, the Riesz representer, and the Neyman orthogonal score are given by
\begin{align*}
m^{\text{ATE}}(W, \gamma_0)&\coloneqq \gamma_0(1,Z) - \gamma_0(0,Z)\\
\alpha^{\text{ATE}}_0(D, Z)&\coloneqq \frac{D}{e_0(Z)}-\frac{1-D}{1-e_0(Z)}.
\end{align*}
Thus, the Neyman orthogonal score is
\begin{align*}
    &\psi^{\text{ATE}}\p{W;\gamma_0, \alpha^{\text{ATE}}_0, \theta^{\text{ATE}}_0}\\
    &\coloneqq\p{\frac{D}{e_0(Z)} - \frac{1 - D}{1-e_0(Z)}}\bigp{Y-\gamma_0(D,Z)} + \gamma_0(1, Z) - \gamma_0(0, Z) - \theta^{\text{ATE}}_0.
\end{align*}
Estimators using such scores are also called doubly robust estimators \citep{Bang2005doublyrobust,Olea2024doublerobustness}. 

\section{Equivalence between Riesz regression and LSIF}
\label{sec:main2}
\subsection{Riesz Regression}
\label{sec:riesz}
Riesz regression \citep{Chernozhukov2021automaticdebiased} estimates the unknown Riesz representer by minimizing a mean squared error criterion. 

\paragraph{General formulation}
Let $\calA$ be a model of $\alpha_0$, such as a linear model, an RKHS class, or a neural network. For $\alpha \in \calA$, define
\[
Q(\alpha)\coloneqq \bbE\sqb{\bigp{\alpha_0(D,Z) - \alpha(D,Z)}^2}.
\]
Riesz regression estimates $\alpha_0$ by minimizing an empirical version of this population risk. Although $Q(\alpha)$ contains the unknown $\alpha_0$, an equivalent feasible objective is available.

The appeal of this formulation is that it targets the Riesz representer itself, rather than estimating the propensity score first and then transforming it into weights. This is useful when one wants to regularize the weights directly or when the representer has a structure that is easier to exploit than a specific propensity score model. The feasible objective shows that direct representer fitting is possible even though $\alpha_0$ is unobserved.

\paragraph{ATE estimation}
In the ATE case,
\begin{align*}
    \alpha^* &\coloneqq \argmin_{\alpha \in \calA} \bbE\sqb{\bigp{\alpha^{\text{ATE}}_0(D,Z) - \alpha(D,Z)}^2}\\
    &= \argmin_{\alpha \in \calA} \bbE\sqb{-2\bigp{\alpha(1,Z) - \alpha(0,Z)} + \alpha(D,Z)^2}.
\end{align*}
The identity follows from
\[
\bbE\sqb{\alpha^{\text{ATE}}_0(D,Z)\alpha(D,Z)}
= \bbE\sqb{\alpha(1,Z)}-\bbE\sqb{\alpha(0,Z)}.
\]
Hence, the empirical estimator is
\[
\widehat{\alpha}\in\argmin_{\alpha\in\calA}
\cb{\frac{1}{n}\sum^n_{i=1}\Bigp{-2\bigp{\alpha(1,Z_i) - \alpha(0,Z_i)} + \alpha(D_i,Z_i)^2}
+\lambda \Omega(\alpha)},
\]
where $\Omega$ is a regularizer, such as the $\ell_2$ norm or the RKHS norm.

\subsection{Direct DRE}
\label{sec:dre}
Direct DRE targets the ratio itself rather than estimating two densities separately \citep{Sugiyama2011densityratio}. Let $X^{(\text{de})}$ follow a distribution with density $p_{\text{de}}$, and let $X^{(\text{nu})}$ follow a distribution with density $p_{\text{nu}}$. Let $\cb{X_j^{(\text{de})}}_{j=1}^{n_{\text{de}}}$ and $\cb{X_k^{(\text{nu})}}_{k=1}^{n_{\text{nu}}}$ be two independent samples, where $X_j^{(\text{de})}$ is an i.i.d. copy of $X^{(\text{de})}\sim p_{\text{de}}$, and $X_k^{(\text{nu})}$ is an i.i.d. copy of $X^{(\text{nu})}\sim p_{\text{nu}}$. Our goal is to estimate
\[
r_0(x) \coloneqq \frac{p_{\text{nu}}(x)}{p_{\text{de}}(x)}.
\]
A straightforward approach estimates $p_{\text{nu}}$ and $p_{\text{de}}$ separately and then takes their ratio. Direct methods target the ratio itself, which avoids error amplification and allows the loss to be matched to the application. This literature includes moment matching, classification-based methods, divergence minimization, and least-squares approaches \citep{Huang2007correctingsample,Qin1998inferencesfor,Nguyen2010estimatingdivergence,Sugiyama2011densityratio}. 

LSIF estimates the density ratio by minimizing
\begin{align*}
    r^* &\coloneqq \argmin_{r \in \calR} \bbE_{p_{\text{de}}}\sqb{\bigp{r_0(X) - r(X)}^2}\\
&= \argmin_{r \in \calR} \cb{-2 \bbE_{p_{\text{nu}}}\sqb{r(X)} + \bbE_{p_{\text{de}}}\sqb{r(X)^2}},
\end{align*}
where $\calR$ is a model of $r_0$. The feasible objective follows from the identity
\[
\bbE_{p_{\text{de}}}\sqb{r_0(X)r(X)} = \bbE_{p_{\text{nu}}}\sqb{r(X)}.
\]
The empirical estimator is
\[
\widehat{r}\in\argmin_{r\in\calR}
\cb{-\frac{2}{n_{\text{nu}}}\sum^{n_{\text{nu}}}_{k=1}r\p{X_k^{(\text{nu})}} + \frac{1}{n_{\text{de}}}\sum^{n_{\text{de}}}_{j=1}r\p{X_j^{(\text{de})}}^2
+\lambda \Omega(r)}.
\]

\subsection{Equivalence}
\label{sec:equivrieszdr}
The equivalence between Riesz regression and LSIF is immediate once the ATE representer is written as two treatment-specific density ratios. Since
\[
e_0(z) = \frac{p_{D,Z}(1,z)}{p_Z(z)},
\]
define
\begin{align*}
r_0(d,z)&\coloneqq \frac{p_Z(z)}{p_{D, Z}(d, z)},
\end{align*}
for $d \in \{1, 0\}$. 
Then
\[
\alpha^{\text{ATE}}_0(D,Z) = Dr_0(1, Z) - (1-D)r_0(0, Z).
\]
Equivalently, $r_0(1,z)=1/e_0(z)$ and $r_0(0,z)=1/\p{1-e_0(z)}$, so the decomposition simply separates the treated and control components of the representer.

For $r_0(1,z)$, the corresponding LSIF criterion can be written as
\begin{align*}
    r^*(1, \cdot) &\coloneqq \argmin_{r(1, \cdot) \in \calR} \int \bigp{r_0(1,z) - r(1,z)}^2 p_{D,Z}(1,z)\rmd z\\
&= \argmin_{r(1, \cdot) \in \calR} \cb{\bbE_{p_Z}\sqb{-2 r(1,Z) + e_0(Z)r(1,Z)^2}}.
\end{align*}
Similarly,
\begin{align*}
    r^*(0, \cdot) &\coloneqq \argmin_{r(0, \cdot) \in \calR} \int \bigp{r_0(0,z) - r(0,z)}^2 p_{D,Z}(0,z)\rmd z\\
&= \argmin_{r(0, \cdot) \in \calR} \cb{\bbE_{p_Z}\sqb{-2 r(0,Z) + \bigp{1-e_0(Z)}r(0,Z)^2}}.
\end{align*}
Combining the two objectives gives
\[
\argmin_{r(1, \cdot), r(0, \cdot) \in \calR}
\cb{\bbE_{p_{D,Z}}\sqb{-2 \bigp{r(1,Z) + r(0,Z)} + Dr(1,Z)^2 + (1-D)r(0,Z)^2}}.
\]
With the substitution
\[
\alpha(D,Z)=Dr(1, Z)-(1-D)r(0, Z),
\]
we have $\alpha(1,Z)=r(1,Z)$, $\alpha(0,Z)=-r(0,Z)$, and $\alpha(D,Z)^2=Dr(1,Z)^2+(1-D)r(0,Z)^2$. Therefore, the LSIF objective becomes
\[
\bbE\sqb{-2\bigp{\alpha(1,Z)-\alpha(0,Z)} + \alpha(D,Z)^2},
\]
which is exactly the ATE Riesz regression objective. This equivalence is exact at the level of the population risk. The empirical Riesz regression criterion is the corresponding plug-in estimator of this population objective under the observed data sampling scheme. Note that Riesz regression can be understood as a joint density ratio estimator that optionally shares basis functions or network parameters across treatment arms. Such shared structure is often attractive in practice because both ratios are learned on the same covariate space and usually benefit from common features.

\section{Implications for Causal Weighting and Regularization}
The equivalence is useful because the representer is the object that generates the orthogonal score weights \citep{Kato2026unifiedframework}. Different density ratio losses are therefore different ways of estimating the same weighting function. Quadratic loss gives LSIF and Riesz regression, the corresponding dual problem yields stable balancing weights \citep{Zubizarreta2015stableweights}. More generally, DRE often replaces squared error with Bregman divergence criteria \citep{Sugiyama2011densityratio}. In the present setting, this yields Bregman-type Riesz regression \citep{Kato2025directbias}. Kullback-Leibler (KL)-type losses connect the density ratio literature to the tailored loss of \citet{Zhao2019covariatebalancing}, and the corresponding dual problem yields entropy balancing weights \citep{Hainmueller2012entropybalancing}. Because the ATE representer is signed, KL-type implementations use the signed modifications developed in \citet{Kato2025directbias}.

The identity also transfers model choice and regularization results. In RKHS models, KuLSIF provides analytic solutions and efficient cross-validation \citep{Kanamori2012statisticalanalysis}. In neural network models deep density ratio methods have been proposed with finite-sample error bounds and nonparametric rates under smoothness assumptions \citep{Kato2021nonnegativebregman,Zheng2022anerror}. When the density ratio is difficult to estimate, the implied representer is unstable, so extreme orthogonal score weights should be expected. This suggests a concrete empirical workflow. One chooses the model class and regularization by thinking directly about the complexity of the weights required by the target parameter, rather than about an auxiliary propensity score model. The DRE literature studies this problem directly. Nonnegative Bregman corrections curb training loss hacking in flexible models, and telescoping DRE replaces one difficult ratio with a product of easier local ratios \citep{Kato2021nonnegativebregman,Rhodes2020telescopingdensityratio}.

The identity also generalizes matching methods. \citet{Lin2023estimationbased} shows that nearest neighbor matching implicitly estimates a density ratio, and \citet{Kato2025nearestneighbor} shows that it can be written as LSIF, hence as Riesz regression, with a matching basis. Under the present view, matching, balancing, and augmentation are not separate ideas. For more detailed arguments about this perspective, see \citet{Kato2026unifiedframework}.

\section{Conclusion}
This paper shows that the key link between Riesz regression and direct DRE is the representation of the Riesz representer as a signed density ratio. This perspective gives a direct weighting interpretation of representer fitting and clarifies the connections among orthogonal scores, balancing weights, matching, augmentation, model selection, and regularization. It turns an abstract Riesz representer into a familiar reweighting object and shows when methods from DRE can be used without changing the target parameter.

\clearpage 

\bibliography{arXiv2.bbl}

\end{document}